# An Evolutionary Approach for Optimal Citing and Sizing of Micro-Grid in Radial Distribution Systems

J.Eswari1, Dr.S.Jeyadevi2
*PG scholar, Professor, Dept of EEE*
*A Kamaraj college of Engineering and Technology,*
*Virudhunagar, India*

*Abstract*— This Paper presents the methodology of penetration of Micro-Grids (MG) in the radial distribution system (RDS). The aim of this paper is to minimize a total real power loss that descends the performance of the radial distribution system by integrating various renewable resources as Distributed Generation (DG). The combination of different types of renewable energy resources contributes a sustainable MG. These resources are optimally sized and located using evolutionary approach in various penetration levels. The optimal solutions are experimented with IEEE 33 radial distribution system using Particle Swarm Optimization (PSO) technique. The results are quite promising and authenticate its potential to solve problem in radial distribution system effectively.

## I. INTRODUCTION

In Recent years, Micro-Grids become prominent and preferable in demand side. The MG consists of micro sources of various renewable resources. The Micro-sources present are of smaller capacity indeed supplying electrical and heating loads close to the customer. It is widely used in the remote areas where the transmission lines are economically not feasible with the maximum capacity of 10 MVA as per IEEE standards. The control techniques for the micro sources enhance the load voltage profile and frequency [1]. The integration of renewable sources in to the RDS improves the performance of the system in all beneficial aspects [2]. The MG is modeled dynamically so that the control over the system to achieve the objectives.

The MG can be DC/AC type and it depends on the utilization. It can also be formed with hybrid resources for optimal energy management and control in [3]. The integration of DG resources to the power system is important to develop the quality of power supply, reliability and performance of the system [4]. The interconnection of micro sources adopts control and synchronizing techniques for better concert. The radial distribution network involves the growth with micro sources has greatest impact in voltage profile, load losses, harmonics and fault recovery. The integration of DG enhances better results in the distribution network only when it is optimally cited and sized [5]. Various methodologies like Genetic Algorithm (GA), PSO, Linear Programming (LP), Hybrid GAPSO used for optimal placement of DGs.

The DGs are modeled and connected to the RDS at Point of Common Coupling (PCC) as real and reactive power components [6]. In [7], the coordination of multiple DGs for real and reactive power management in Micro-Grids are implemented. The Power management strategy during small signal dynamics is performed in grid connected mode, autonomous mode and intermediate stage between the two modes.

The radial Distribution Network has the structure which emerges from the root node where the generation is connected. The RDS has root node or main node followed by lateral line. The lateral line emerges from the main feeder and connecting loads. The sub-lateral line emerges from the lateral line. Finally the minor line emerges from the sub-lateral line. Due to the radial structure, high R/X ratio, the analysis of RDS becomes quiet complicated and hence conventional method like Newton Raphson (NR) is not always suitable. Methods like ladder type, Branch current calculation, kirchoff's laws and forward and backward sweep algorithm are mostly preferable [8].

The loads connected to RDS may be of constant power load, constant current and constant impedance type [9]. The loads can be a voltage dependant by modeling static polynomial load. Different types of static loads with various tolerance levels promote the load flow solutions of the network. Polynomial equations used for analysis provide better convergence results in the RDS.

The load flow solution for the RDS can be calculated by different methods. Branch flow equation method with the break points at some buses results in less convergence time [10]. The Backward-Forward Sweep (BFS) algorithm has better performance in RDS load flow [11]. The improved version of BFS algorithm with the BIBC matrix manipulation provided better voltage profile results [12].

This paper aims at minimizing the real power losses in the RDS by integrating the MG with it. A swarm intelligence technique is used to locate and size the micro-sources in the MG. The MG is integrated in to the network in various penetration levels. This paper contains six sections. Section II presents the problem formulation and objective if this paper. Section III describes the citing and sizing. Section IV describes the proposed algorithm technique PSO to solve the problem formulated. Section V discusses about the results and finally Section VI concludes the paper.

## II. PROBLEM FORMULATION

*A. Objective Function*

The aim of this paper concentrates on minimization of real power losses in the radial network which is given in Eqn. (1). The objective function is subjected to equality and inequality constraints. The equality constraints considering the total sized DG units should be equal to the DG penetration level as shown in (2). The inequality constraint involves the generator rating, Voltage constraint and the branch current constraint.

*Minimize*

$$F = \sum_{i=1}^{N} \sum_{k=1}^{N} P_{loss_{ik}} \quad (1)$$

*Subject to*
*Equality Constraint*

$$PDG_1 + PDG_2 + ... + PDG_r = PDG_{pp} \quad (2)$$

*Inequality Constraint*

Generator Rating Constraint

$$P_{gi\,min} \leq P_{gi} \leq P_{gi\,max} \quad (3)$$

Voltage Constraint

$$V_{min} \leq V_i \leq V_{max} \quad (4)$$

Branch Current Constraint

$$I_{(i,j)} \leq I_{rated} \quad (5)$$

Where $P_{loss_{ik}}$ is the real power losses from the bus i to bus k

$N$ is the total number of buses in the system
$PDG$ is the real power generation capacity of DG
$r$ is the number of DG
$PDG_{pp}$ is the maximum Real power penetration of DG

$P_{gi\,min}$ is the Real Power minimum bounds set for the of $i^{th}$ generator

$P_{gi\,max}$ is the Real Power maximum bounds set for the of $i^{th}$ generator

$P_{gi}$ is the real power rating of the $i^{th}$ generator.

$V_{min}$ is the minimum Voltage magnitude on the bus

$V_{mac}$ is the maximum Voltage magnitude on the bus

$V_i$ is the Voltage magnitude on the bus i

$I_{(i,j)}$ is the feeder current in the branch ij

$I_{rated}$ is the rated current of the feeder

## III. CITING AND SIZING

The Micro-Grid which consists of multiple resource supplying its own heat and electrical loads is integrated in to the RDS is shown in Fig.1. There are many challenges present in integration of renewable energy sources. The Micro-grid considered consists of three different types of resources such as PV, Wind and Micro-turbine. These micro sources are falls under different type of DG category that inject or absorb real and reactive powers. In this paper the loads existing in the chosen RDN system is supplied with the micro sources integrated at optimal location.

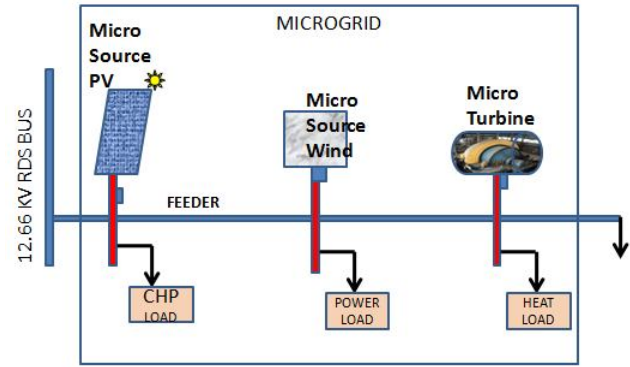

Fig.1 Micro-Grid Structure integrated with RDS

A. *Citing Of Micro-Grid In RDS*
*1) Radial Network Configuration*

In general, the RDS is radial in nature and it does not have a loop structure. The general configuration of the radial structure consists of a root node where the generator is connected through the distribution transformer. The source is available only at starting point of the RDS. The nodes are numbered in ascending order. Every adjacent nodes are connected by branches which are number uniquely. A sample radial distribution network is shown in the Fig.2 [13].

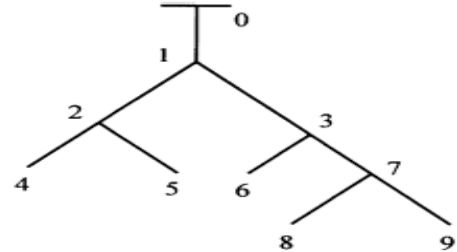

Fig.2 A Typical Radial Distribution network

*2) Load flow method*

There are several types of load flow calculation methods available for RDS. The conventional type NR method is not suitable due to the high R/X ratio that causes poor convergence characteristics. In this paper, the Backward and Forward Sweep algorithm is implemented due to its improved characteristics.

*3) Citing Algorithm*

The Micro sources can be optimally placed by calculating the weak bus in the system. The weak bus can be identified by Maximum loadability index (MLI). It provides the P and Q value at the maximum loadability margin beyond which the system becomes unstable. It is given as follows in (6)

$$MLI = \frac{|V_p|^2 \left[ -(r_{ij}P_j + x_{ij}Q_j) + \sqrt{(r^2_{ij} + x^2_{ij})(P^2_j + Q^2_j)} \right]}{2\left[ \left(\sqrt{(r^2_{ij} + x^2_{ij})(P^2_j + Q^2_j)}\right)^2 - (r_{ij}P_j + x_{ij}Q_j)^2 \right]} \quad (6)$$

Based on the loadability index obtained, the candidate buses are selected for penetrating the micro-sources in the RDS.

*B. Sizing of Micro-Grid*

The sizing of the MG in RDS can be performed using analytical method as well as intelligence techniques. The DGs can be modeled in three different types. They are Negative load model, P and Constant PF model and PV model.

*1) Micro-Grid Model*

The micro sources chosen to integrate with RDS must be modeled such that it can be analyzed using analytical/Artificial intelligence methods. In this paper, the micro sources are modeled as P and Constant PF. The power factor varies for different types of DG. The micro sources such as Photo Voltaic (PV) has maximum of PF as 1. Others like a Wind and Micro turbine has PF vary between 0.85-0.9.

*2) Micro-Grid Penetration*

The Penetration level of MG in RDS implies the contribution of micro sources to satisfy the existing demand. The level of penetration can be calculated by the following index as given in (7)

$$PL(\%) = \frac{S_{DG}}{S_{load}} * 100 \quad (7)$$

*3) Analytical Method*

The analytical methods involves sizing of PDG and QDG based up the given expression (8) and (9).

$$P_{DGi} = \frac{\alpha_{ii}(P_{Di} + Q_{Di}) + \beta_{ii}(aP_{Di} - Q_{Di}) - X_i - aY_i}{a^2\alpha_{ii} + \alpha_{ii}} \quad (8)$$

$$Q_{DGi} = aP_{DGi} \quad (9)$$

Where a $=$ (sign) tan (cos$^{-1}$ (PF$_{DG}$))

The sign depends on Power injection/Absorption. The analytical method of sizing DG and injecting power to the RDS has designed the following Micro source DG units of the Micro-grid. Table I presents the loss reduction when the analytically sized DGs are integrated in to the RDS. It is clear that the losses in the IEEE 33 bus RDS test system is reduced 16.56% when designed with analytically sized Micro-grid.

TABLE I ANALYTICAL SIZING OF DG UNITS

| Type | Base Case | Analytical Method | Loss (%) |
|---|---|---|---|
| Real Power Loss (KW) | 169 | 141 | 16.56 % |
| Reactive Loss(KVAr) | 117 | 98 | 16.23% |

## IV. PARTICLE SWARM OPTIMIZATION

The PSO is proposed by Kennedy and Eberhart in the mid 1990s while attempting to simulate graceful motion of swarms of birds investigating the notion of "collective intelligence" in biological species. It is a swarm intelligence algorithm. PSO imitates the flock of birds, schools of fish to find their food source. PSO is a population based algorithm where each individual is identifies as particles [14].

A. PARAMETERS OF PSO

The particle swarm consists of "n" particles. Each particle has its own position X and velocity V. The best position of the particle until the current time with respect to the objective of the system is called personal best (pbest). Among all the particles in the swarm, the best position of the particle with respect to the objective function is called Global best (gbest). c1 and c2 are the acceleration coefficients usually chosen in the range between 0 and 2. The random values r1 and r2 are generated from uniform distribution in the range [0 1]. W is the inertial weight proposed by Eberhart R C,Shi Y,1998 [13]. It is the proportional agent which is responsible for speed control of the algorithm. If the x is smaller, then the searching ability is greater.

The particle in the swarm follows three principles:
  (i) To maintain its inertia
  (ii) To change the condition based on fitness value
  (iii) To change the condition based on swarm's fitness value

*B. Algorithm:*

STEP:1 Initialize the swarm
  (i) Define the inertial weight w, acceleration coefficients c1, c2 values
  (ii) Define the bounds of the velocity $V_{max}$ and $V_{min}$.
  (iii) Assign the initial position $X_i$ and velocity $V_i$
  (iv) $X_i = [x_{i1}, x_{i2}, \ldots, x_{in}]$  $V_i = [v_{i1}, v_{i2}, \ldots, v_{in}]$
  (v) Initialize the population of N particles with D dimension

STEP: 2 Evaluation of fitness function

The fitness function for the swarm involves minimization of real power losses for the selected DG size. The Control variables are tuned to get minimum real power losses.

STEP:3 Update the particle velocity and position using the expression (10) and (11)

$$V_i^{k+1} = wV_i^k + c_1 r_1 (Xp_{best}^k - X_i^k) + c_2 r_2 (Xg_{best}^k - X_i^k) \quad (10)$$

$$X_i^{k+1} = X_i^k + V_i^{k+1} \quad (11)$$

In velocity update equation, the first component is called the inertia component. It enhances the convergence of the problem to optimal solution. The second term $c_1 r_1 (Xp_{best}^k - X_i^k)$ is the cognitive component representing the particle's memory and the third term $c_2 r_2 (Xg_{best}^k - X_i^k)$ is the social component which moves the particle towards the best region.

STEP 4 Evaluate the fitness function for the entire particle after updating the velocity and position.

STEP 5 Evaluate the pbest of the population after updating the history of each particle.

STEP 6 Display the gbest after satisfying the termination criteria. Termination criteria may be

    (i)    Maximum number of iterations
    (ii)    Minimum fitness value

In this paper the numbers of iterations are set to 100. Table.II provides the PSO output parameters and the fitness value obtained for the MG in RDS.

Table II PSO Parameters

| PSO PARAMETERS | ATTRIBUTE VALUES |
|---|---|
| C1 | 1 |
| C2 | 1 |
| Particle Size | [1x6] |
| Population Size | 50 |
| Velocity | ±125 |

The optimal sizing of the micro sources in MG is obtained and it is converged in 52$^{nd}$ iteration as shown in Fig.3.

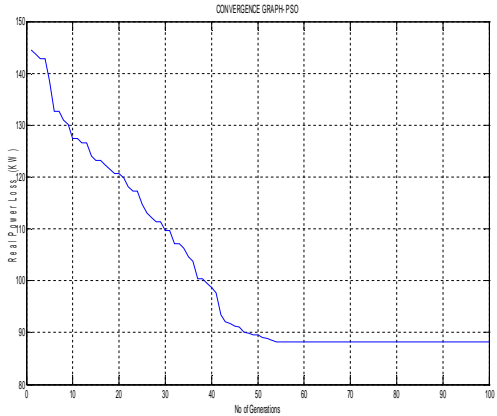

Fig.. Convergence Characteristics of PSO

V RESULTS AND DISCUSSIONS

In this section, the proposed algorithm is applied to achieve the loss minimization objective for various penetration levels of micro sources. The IEEE 33 bus distribution system is considered as the test system. The test system is taken with the base values of 12.66 KV and 100 MVA. The branch and the load details are considered as in [14]. The test system is shown in the Fig. 3. It has a main line and three lateral lines. The main line ends with 18th node. The test system is energized with single source.

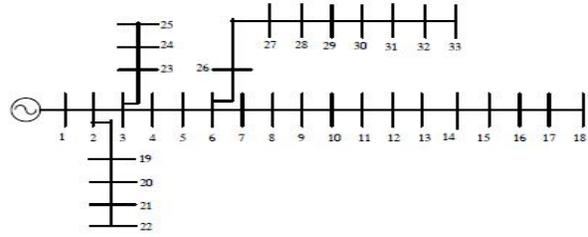

Fig.4. IEEE 33 bus Radial Distribution system

A. *Voltage Profile of the Test System*

The IEEE 33 bus test system is designed with the flat voltage profile for initial load flow analysis. The Backward and Forward sweep based load flow is used for analyzing the test system. It is found that the voltage profile of the system when analyzed in base case and after Micro-Grid integration in Analytical and PSO modes. Table III shows the comparison of improved voltage profile for base case, analytical and PSO.

TABLE III COMPARISON OF VOLTAGE MAGNITUDE IN BASE CASE, ANALYTICAL AND PSO

| Bus No | Voltage Magnitude (P.U) | | |
|---|---|---|---|
| | Base Case | Analytical | PSO |
| 1 | 1 | 1 | 1 |
| 2 | 0.998 | 0.9982 | 0.9976 |
| 3 | 0.9878 | 0.9887 | 0.9854 |
| 4 | 0.9804 | 0.9819 | 0.9766 |
| 5 | 0.9731 | 0.9753 | 0.9679 |
| 6 | 0.9549 | 0.9586 | 0.9496 |
| 7 | 0.9514 | 0.9551 | 0.9478 |
| 8 | 0.9467 | 0.9503 | 0.9424 |
| 9 | 0.9405 | 0.9441 | 0.9366 |
| 10 | 0.9347 | 0.9383 | 0.9313 |
| 11 | 0.9339 | 0.9374 | 0.9304 |
| 12 | 0.9324 | 0.9359 | 0.9289 |
| 13 | 0.9263 | 0.9299 | 0.9239 |
| 14 | 0.9241 | 0.9276 | 0.9224 |
| 15 | 0.9227 | 0.9262 | 0.9216 |
| 16 | 0.9213 | 0.9249 | 0.9209 |
| 17 | 0.9193 | 0.9229 | 0.9191 |
| 8 | 0.9187 | 0.9223 | 0.9184 |
| 19 | 0.9975 | 0.9976 | 0.9973 |
| 20 | 0.9939 | 0.9941 | 0.9955 |
| 21 | 0.9932 | 0.9934 | 0.9953 |
| 22 | 0.9926 | 0.9927 | 0.9947 |
| 23 | 0.9842 | 0.9851 | 0.9814 |
| 24 | 0.9775 | 0.9785 | 0.9742 |

| Bus No | Voltage Magnitude (P.U) | | |
|---|---|---|---|
| | Base Case | Analytical | PSO |
| 25 | 0.9742 | 0.9752 | 0.9706 |
| 26 | 0.953 | 0.957 | 0.947 |
| 27 | 0.9504 | 0.955 | 0.9435 |
| 28 | 0.939 | 0.9457 | 0.9311 |
| 29 | 0.9309 | 0.9392 | 0.9222 |
| 30 | 0.9273 | 0.9365 | 0.9172 |
| 31 | 0.9232 | 0.9344 | 0.913 |
| 32 | 0.9223 | 0.9341 | 0.9121 |
| 33 | 0.922 | 0.9343 | 0.9119 |

The Voltage profile of the sizing methodology of PSO is improved when compared to analytical and base case. It is shown in Fig.5.

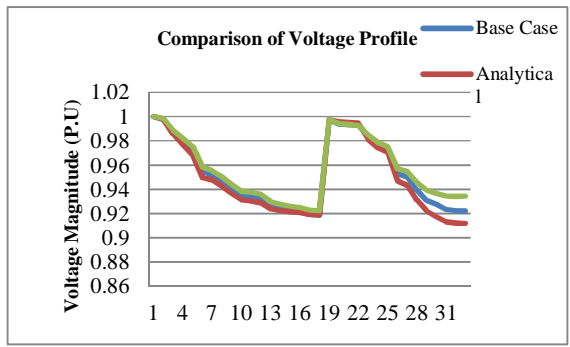

Fig.5. Comparison of Voltage Profile

B. *Micro-Grid Penetration*

The Sizing of Micro-grid with various penetration levels is implemented. Based on various penetration levels, the objective function is optimally achieved. Fig.6 shows the minimization of real power losses using PSO with different penetration levels say 50%, 60% and 80 % respectively.

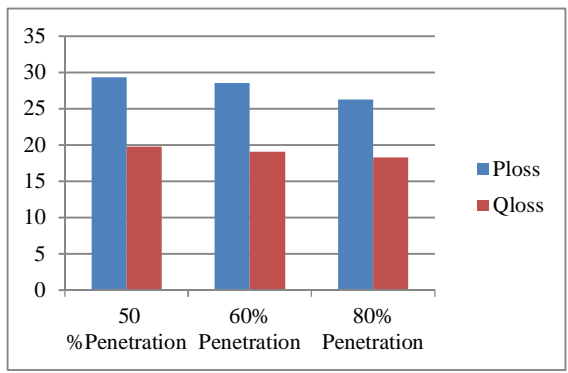

Fig.6 MG Penetration Percentage variation in RDS

## VI CONCLUSION

This paper develops an evolutionary methodology to solve the loss minimization in RDS considering penetration of MG in to account. Citing of MG resources by MLI has improved the design feature of the radial distribution network. The sizing solution provided by PSO algorithm gives better results of real power losses. The swarm intelligence technique parameter pbest and gbest enhances the objective by tuning it to an optimal level.It also enhances renewable energy penetration that enhances the performance of the RDS. The significance of the MG integration in RDS improves the reduction of losses up to 26.22KW with the maximum of 80% of penetration.